\documentclass[runningheads]{llncs}
\usepackage[T1]{fontenc}
\usepackage{graphicx}
\usepackage{graphicx}
\usepackage{multirow,multicol}
\usepackage{adjustbox}
\usepackage{amsmath}
\usepackage{booktabs}
\usepackage{amssymb}
\usepackage{mathtools}
\usepackage{nicefrac}
\usepackage{graphicx}
\usepackage{float}
\usepackage[hyphens]{url}
\usepackage{algorithm}
\usepackage{algorithmic}
\usepackage{graphicx}
\usepackage{graphicx}
\usepackage{subcaption}
\usepackage{xspace}
\usepackage{array}
\usepackage{adjustbox}
\usepackage{microtype}
\usepackage{ntheorem}
\usepackage{adjustbox}
\usepackage{bm}
\usepackage[misc,geometry]{ifsym}
\begin{document}
%
%\title{Enhancing Forecasting Robustness: Time Series Clustering for Stress Testing and Debugging}

%\title{Towards Robust Forecasting Performance Evaluation and  Adaptive Prediction Intervals Estimation}

%\title{Bridging Model Evaluation and Adaptive Prediction Intervals Estimation for Forecasting}
\title{Adaptive Fine-Tuning via Pattern Specialization for Deep Time Series Forecasting \thanks{This research has partly been funded by the Federal Ministry of Education and Research of Germany and the state of North-Rhine-Westphalia as
part of the Lamarr-Institute for Machine Learning and Artificial Intelligence.}}
\titlerunning{Adaptive Fine-Tuning for Deep Time Series Forecasting}
\author{Amal Saadallah \orcidID{0000-0003-2976-7574} \Letter 
\and
Abdulaziz Al-Ademi}
\authorrunning{Saadallah and Al-Ademi}
\institute{Lamarr Institute for Machine Learning and AI, Dortmund, Germany \\
\email{\Letter amal.saadallah@cs.tu-dortmund.de}}
\maketitle              % typeset the header of the contribution
\begin{abstract}

Time series forecasting poses significant challenges in non-stationary environments where underlying patterns evolve over time. In this work, we propose a novel framework that enhances deep neural network (DNN) performance by leveraging specialized model adaptation and selection. Initially, a base DNN is trained offline on historical time series data. A reserved validation subset is then segmented to extract and cluster the most dominant patterns within the series, thereby identifying distinct regimes. For each identified cluster, the base DNN is fine-tuned to produce a specialized version that captures unique pattern characteristics. At inference, the most recent input is matched against the cluster centroids, and the corresponding fine-tuned version is deployed based on the closest similarity measure. Additionally, our approach integrates a concept drift detection mechanism to identify and adapt to emerging patterns caused by non-stationary behavior. The proposed framework is generalizable across various DNN architectures and has demonstrated significant performance gains on both traditional DNNs and recent advanced architectures implemented in the GluonTS library.

\end{abstract}

\section{Introduction}

Time series forecasting is a fundamental task that underpins a wide range of real-world applications, including finance, energy management, healthcare, and supply chain optimization \cite{hyndman2018forecasting,cerqueira2017arbitrated,saadallah2021explainable,saadallah2020drift}. Accurate and reliable forecasts are essential for enabling informed decision-making, optimizing resource allocation, and ensuring the smooth operation of complex systems. However, delivering consistent predictive accuracy in practice remains a big challenge due to the inherent complexity, non-stationarity, and dynamic behavior of temporal data \cite{saadallah2020drift,saadallah2023online,saadallah2023ecml}.

A major challenge in time series forecasting is the lack of a single, universally effective model that can generalize across different application domains, datasets, and temporal horizons \cite{cerqueira2017arbitrated,cerqueira2019arbitrage}. Forecasting models that excel under specific conditions or during stable periods often fail to sustain their performance when confronted with evolving patterns and distributional shifts—phenomena that are common in real-world time series data \cite{saadallah2021explainable,saadallah2020drift,saadallah2018bright}. These shifts, often referred to as \emph{concept drift}, can cause a substantial deterioration in predictive accuracy if not properly detected and mitigated \cite{saadallah2020drift}.
%A fundamental challenge in time series forecasting lies in the absence of a single, universally optimal model capable of maintaining consistently high performance across varying application domains, datasets, and temporal horizons \cite{cerqueira2017arbitrated}. More critically, even within the same time series, different temporal segments often exhibit distinct patterns and statistical properties—such as varying trends, seasonality, or volatility—that can significantly influence a model’s predictive accuracy \cite{saadallah2021explainable,saadallah2022explainable,saadallah2023online}. As a result, a model that performs well during one regime or pattern may fail to generalize effectively when the underlying data distribution shifts \cite{saadallah2020drift}. This time-varying behavior underscores the need for adaptive forecasting approaches that can dynamically specialize or select models based on the local characteristics of the time series data \cite{saadallah2021explainable,saadallah2023online}. 

In addition to global distributional changes, time series data often exhibit localized variations and recurring patterns that can significantly affect model performance \cite{saadallah2021explainable,saadallah2022explainable,saadallah2023online}. Different forecasting models may specialize in capturing distinct types of temporal behavior, such as trends, seasonality, or abrupt transitions. Consequently, a model that performs well in one segment of the data may underperform in another due to its inability to adapt to localized characteristics \cite{saadallah2021explainable}. This highlights the critical need for adaptive forecasting frameworks that can (i) detect and respond to concept drift, (ii) capture local temporal dynamics, and (iii) leverage model specialization to improve robustness and predictive accuracy.

Deep neural networks (DNNs) have recently demonstrated significant success in time series forecasting tasks due to their ability to model complex nonlinear dependencies and capture long-range temporal patterns \cite{livieris2020cnn,gers2002applying}. Advanced architectures, such as Temporal Fusion Transformers (TFT) \cite{lim2021temporal}, DeepAR \cite{salinas2020deepar}, and MQ-CNN \cite{wen2017multi}, implemented in libraries like GluonTS, offer scalable and flexible solutions for probabilistic forecasting. Despite their impressive performance, DNN-based models often operate under the assumption of stationarity or require frequent retraining to adapt to new data distributions. Furthermore, these models typically learn a global mapping from inputs to outputs, which can hinder their ability to generalize across distinct regimes or localized patterns within the same time series.
Moreover, while DNNs excel at capturing complex structures when abundant training data is available, their generalization capability can be compromised in the presence of concept drift and heterogeneous data segments \cite{saadallah2021explainable,saadallah2022explainable}. Static DNN models lack inherent mechanisms to adapt rapidly to evolving temporal patterns, leading to performance degradation over time unless retrained or explicitly augmented with adaptive components \cite{saadallah2022explainable}.

%\subsection*{Our Proposal and Contributions}

In this paper, we propose a novel adaptive pattern-based framework for time series forecasting that addresses the limitations of conventional DNN approaches. Our framework leverages the notion of pattern specialization by clustering historical time series data to identify dominant temporal regimes. For each cluster, we fine-tune a specialized version of a base DNN, thereby creating an ensemble of expert DNNs, each tailored to a specific data regime. During inference, the most recent input subsequence is compared to the cluster centroids, and the closest matching expert DNN is selected for prediction. This process enables dynamic model selection that aligns with the current temporal context, improving forecasting accuracy and robustness.
To further enhance adaptability, we incorporate a concept drift detection mechanism that identifies emerging patterns in the time series data. When a new pattern is detected that does not align with existing clusters, the framework triggers the creation of a new specialized DNN or updates the existing model pool. This ensures that the forecasting system remains responsive to non-stationary dynamics without requiring full retraining.

The main contributions of this work are summarized as follows:
\begin{itemize}
    \item We propose an adaptive pattern-based framework for time series forecasting that leverages clustering and local specialization to address data heterogeneity and concept drift.
    \item We demonstrate the generalization of the framework across various DNN architectures such as CNN, LSTM, TFT, DeepAR, and MQ-CNN.
    \item We integrate a concept drift detection mechanism to maintain up-to-date model specialization and ensure robustness against non-stationary behavior.
    \item We conduct extensive experiments on diverse time series datasets to evaluate the effectiveness of our approach. Results show consistent improvements in predictive accuracy over traditional DNN training.
\end{itemize}

\section{Related Work}
\label{sec:related_work}
%Time series forecasting has seen significant advancements with the rise of deep neural networks (DNNs), which have demonstrated state-of-the-art performance in capturing complex temporal dependencies and non-linear patterns \cite{livieris2020cnn,gers2002applying,mlp}. 
In this section, we review existing literature on deep learning-based forecasting models, as well as approaches aimed at model specialization and adaptation for handling pattern heterogeneity and non-stationarity in time series data.

\subsection{On Deep Learning for Time Series Forecasting}

Recent advancements in deep learning have led to a proliferation of neural network architectures tailored for time series forecasting. Recurrent neural networks (RNNs), particularly long short-term memory (LSTM) networks \cite{gers2002applying}, have been extensively adopted due to their ability to capture long-range temporal dependencies. In parallel, convolutional neural networks (CNNs) have demonstrated competitive performance in forecasting tasks, offering improved computational efficiency compared to RNN-based models, particularly when handling high-dimensional time series data \cite{livieris2020cnn}.

Comprehensive toolkits such as GluonTS \cite{alexandrov2020gluonts} have further accelerated the adoption of deep neural networks (DNNs) by providing standardized implementations of various state-of-the-art architectures. 
%Notable examples include DeepAR, which leverages RNNs with LSTM or gated recurrent units (GRUs) to produce probabilistic forecasts \cite{salinas2020deepar}, and DeepState, which integrates RNN-based temporal modeling with Kalman filtering to capture hierarchical dependencies across multiple time series \cite{rangapuram2018deep}. Additionally, models such as MQ-RNN and MQ-CNN combine recurrent or convolutional encoders with quantile-based decoders to perform sequence-to-sequence forecasting with uncertainty estimation \cite{wen2017multi}. DeepFactor advances this line of work by jointly modeling global factors and local patterns to improve the accuracy of multi-series forecasting \cite{wang2019deep}.
Notable examples include DeepAR, which utilizes autoregressive recurrent neural networks (RNNs), specifically LSTMs or gated recurrent units (GRUs), to generate probabilistic forecasts \cite{salinas2020deepar}. DeepAR has demonstrated strong performance in capturing complex temporal dependencies through sequential modeling. Similarly, DeepState extends RNN-based modeling by incorporating Kalman filtering to improve probabilistic forecasting  \cite{rangapuram2018deep}. Other architectures, such as MQ-RNN and MQ-CNN, focus on sequence-to-sequence forecasting by integrating recurrent or convolutional encoders with quantile-based decoders to estimate prediction intervals and provide uncertainty quantification \cite{wen2017multi}. DeepFactor advances this line of work by jointly modeling global factors and local patterns \cite{wang2019deep}.
More recently, transformer-based architectures have emerged as powerful alternatives for time series forecasting, owing to their capacity to model complex dependencies without relying on recurrent structures \cite{lim2021time}. These models, inspired by the success of transformers in natural language processing, offer improved scalability and flexibility, particularly in settings involving long-range temporal correlations.

\subsection{On Model Specialization and Adaptation Across Patterns}

Despite the advancements in DNN architectures, a fundamental challenge remains: achieving robust forecasting performance across varying data regimes and under non-stationary conditions. Most existing DNN models are designed to learn a global mapping from historical observations to future predictions, limiting their ability to adapt to evolving patterns or heterogeneous temporal behaviors present in real-world data.
To address these limitations, recent works have explored adaptive and specialized model selection strategies. Saadallah et al. \cite{saadallah2023online} proposed an online deep hybrid ensemble learning framework that dynamically selects and weights deep neural networks based on their local performance in streaming time series data. This approach enables the system to adapt to concept drift and heterogeneous temporal regimes through an ensemble of specialized models.
In a related line of work, in \cite{saadallah2021explainable}, an explainable online deep neural network selection framework that leverages adaptive saliency maps to guide the selection of the most relevant DNN model for each input pattern is introduced. This method not only enhances forecasting accuracy by exploiting localized model expertise but also improves interpretability by highlighting the input time series segments influencing model selection.
Further advancing the concept of model specialization, an explainable online ensemble of DNNs pruning approach is proposed in \cite{saadallah2022explainable}. By dynamically pruning and combining deep neural networks, their method adapts the ensemble structure to reflect the evolving relevance of different models across changing input patterns, maintaining both predictive accuracy and computational efficiency.

%\subsection{Positioning of Our Work}

While prior approaches have demonstrated the benefits of adaptive deep neural network (DNN) selection and pruning to address concept drift and data heterogeneity \cite{saadallah2021explainable,saadallah2022explainable,saadallah2023online}, they primarily rely on online learning mechanisms and ensemble strategies that dynamically adjust model selection or weighting at inference time. These methods often focus on switching between independently trained architectures or pruning weaker models from the ensemble without explicitly improving the training processes or convergence of the individual DNNs themselves. Specifically, they rely on traditional training paradigms in which each model is optimized on the entire historical time series, failing to account for localized temporal patterns or dominant regimes that may benefit from targeted specialization. As a result, while these approaches can enhance predictive accuracy through selection and aggregation, they do not directly address the underlying challenge of adapting DNN parameters to distinct time-varying behaviors within the data. In contrast, our proposed framework introduces a novel offline pattern-based model specialization strategy, where a base DNN is fine-tuned on clustered dominant patterns to create a set of specialized experts. This design allows for tailored model adaptation to distinct regimes identified during training, with efficient model selection based on similarity to recent input patterns during inference. Additionally, our integration of a concept drift detection mechanism enables the dynamic inclusion of new expert models as emerging patterns are observed, bridging the gap between offline specialization and online adaptation.

%By building on the strengths of prior model specialization and adaptation methods \cite{saadallah2021explainable,saadallah2022explainable,saadallah2023online}, our approach provides a generalizable framework that can be applied to various deep learning architectures, including state-of-the-art models implemented in GluonTS, such as TFT and DeepAR. This work extends the existing body of research by systematically addressing local performance optimization and enhancing the robustness of forecasting in non-stationary environments.

\section{Methodology}
\label{sec:methodology}
\subsection{Notation and Problem Formulation}
% We focus on univariate time series forecasting, where the goal is to predict the next value(s) of a time series given its past observations. Let $\mathcal{T} = \{x_t\}_{t=1}^N$ be a time series of length $N$. We consider a step-wise forecasting strategy, where at each time $t$, a model forecasts the value $x_{t+1}$ based on the preceding $p$ observations $S_{t - p + 1:t} = \{x_{t - p + 1}, \dots, x_t\}$.
% The dataset is split into three disjoint subsets:
% \begin{itemize}
%     \item \textbf{Training set} $\mathcal{T}_{\text{train}}$: Used to train the base forecasting model on the entire historical time series data.
%     \item \textbf{Validation set} $\mathcal{T}_{\text{val}}$: Used to extract and cluster the dominant patterns to create specialized models.
%     \item \textbf{Test set} $\mathcal{T}_{\text{test}}$: Used for inference and performance evaluation.
% \end{itemize}

We address the problem of univariate time series forecasting, where the objective is to predict future values of a time series based on its historical observations. Formally, let $\mathcal{T} = \{x_t\}_{t=1}^N$ represent a univariate time series of length $N$, where $x_t \in \mathbb{R}$ represents the value of $\mathcal{T}$ at time step $t$. The forecasting task involves learning a DNN $f(\cdot)$ that maps an input window of the $p$ most recent observations, $S_{t} = \{x_{t - p + 1}, \dots, x_t\}$, to a forecast $\hat{x}_{t+1}$ of the next time step, i.e., $\hat{x}_{t+1} = f(S_{t})$.

To achieve this, the time series $\mathcal{T}$ is partitioned into three disjoint subsets with distinct roles. The training set, denoted as $\mathcal{T}_{\text{train}}$, is used to train a base deep neural network (DNN) model on the entire historical time series data. This DNN learns a general representation of the underlying temporal dynamics. The validation set, $\mathcal{T}_{\text{val}}$, serves to identify and cluster dominant patterns within the time series. These clusters are used to derive specialized DNNs that capture distinct temporal behaviors present in different regimes. The test set, $\mathcal{T}_{\text{test}}$, is employed for inference and the evaluation of forecasting performance. 

The ultimate goal is to enhance forecasting accuracy and robustness in non-stationary environments by adapting and specializing deep learning models to the dominant patterns present in different segments of the time series. During inference, the framework dynamically selects the most suitable specialized DNN based on the similarity between the current input window and the learned cluster centroids, thereby addressing the limitations of conventional, globally trained models in the presence of concept drift and heterogeneous temporal behaviors.

% \subsection{Dominant Patterns Identification}
% \label{sec:cl}
% We perform clustering over the validation subsequences to identify $K$ dominant patterns. Formally, given the set of subsequences $\mathcal{S}_{\text{val}} = \left\{ S_i \right\}_{i=1}^{\lfloor \frac{|\mathcal{T}_{\text{val}}|}{p} \rfloor},$
% we apply $K$-Means clustering with the Euclidean distance metric: $d(S_i, S_j) = \| S_i - S_j \|_2$, where $S_i$ and $S_j$ are subsequences of length $p$. 
% The primary objective of the clustering is to group subsequences that exhibit similar aligned temporal patterns. Notably, the chosen subsequence length \(p\) corresponds to the input size utilized during initial DNN training. 
% The clustering procedure results in $K$ clusters $\mathcal{C} = \left\{ C_k \right\}_{k=1}^{K}$,  where each cluster $C_k$ represents a dominant pattern in the data. The centroid $\mu_k$ of each cluster is computed as:
% $\mu_k = \frac{1}{|C_k|} \sum_{S_i \in C_k} S_i$
% and serves as a reference for determining the similarity of future input sequences during inference.

% For each cluster $C_k$, we fine-tune the base DNN $M_{\text{base}}$ on the corresponding validation subsequences $\left\{ S_i \in C_k \right\}$, yielding a specialized DNN $M_k$. This fine-tuning process enables each $M_k$ to capture the specific dynamics of its associated dominant pattern, thereby enhancing forecasting accuracy in regions of the time series that exhibit similar behaviors.

\subsection{Dominant Patterns Identification}
\label{sec:cl}

To capture the diverse temporal dynamics within the time series data, we perform clustering over the validation subsequences to identify $K$ dominant patterns. Specifically, we partition the validation set $\mathcal{T}_{\text{val}}$ into a collection of non-overlapping subsequences $\mathcal{S}_{\text{val}} = \left\{ S_i \right\}_{i=1}^{\lfloor \frac{|\mathcal{T}_{\text{val}}|}{p} \rfloor}$, where each subsequence $S_i$ is of length $p$. The length $p$ is consistent with the input window size used during the training of the base DNN model $M_{\text{base}}$, ensuring alignment between clustering and model training procedures.
We employ $K$-Means clustering with the Euclidean distance metric, defined as $d(S_i, S_j) = \| S_i - S_j \|_2$, to group subsequences that exhibit similar temporal structures. The primary objective of this clustering process is to uncover regions of the time series that share common temporal characteristics, which we refer to as dominant patterns.

The clustering procedure produces $K$ distinct clusters $\mathcal{C} = \left\{ C_k \right\}_{k=1}^{K}$, where each cluster $C_k$ represents a distinct dominant pattern. The centroid $\mu_k$ of each cluster is computed as: $\mu_k = \frac{1}{|C_k|} \sum_{S_i \in C_k} S_i$.
These centroids serve as representative references and are later used to assign incoming test sequences to the most appropriate specialized model during inference. For each cluster $C_k$, we fine-tune the base DNN $M_{\text{base}}$ on the corresponding validation subsequences $\left\{ S_i \in C_k \right\}$, yielding a specialized DNN $M_k$. This fine-tuning process enables each $M_k$ to capture the specific dynamics of its associated dominant pattern, thereby enhancing forecasting accuracy in regions of the time series that exhibit similar behaviors.
In order to ensure that each cluster $C_k$ contains a sufficient number of subsequences to enable effective fine-tuning of the specialized DNN $M_k$, we impose a size constraint ($n_{\text{min}}$) on the clustering process. 

 \subsection{Specialization via Fine-Tuning}

\subsubsection{Initialization: Base Model Training and Weight Transfer}

The initialization phase of our framework begins with training a general-purpose base DNN $M_{\text{base}}$ on the entire training set $\mathcal{T}_{\text{train}}$. This model is responsible for learning a broad representation of the time series data, capturing the overall dynamics present in the historical observations. $M_{\text{base}}$ serves as the foundation for the specialization process, providing a well-initialized set of weights that will later be adapted to specific dominant patterns.
Formally, the base model is optimized by minimizing the forecasting loss $\mathcal{L}_{\text{forecast}}$ over $\mathcal{T}_{\text{train}}$:
\begin{equation}
\theta_{\text{base}} = \arg\min_{\theta} \mathcal{L}_{\text{forecast}} \left( M_{\theta}(\mathcal{T}_{\text{train}}), \mathcal{T}_{\text{train}}^{\text{target}} \right),
\end{equation}
where $\theta_{\text{base}}$ denotes the learned weights of $M_{\text{base}}$, and $\mathcal{T}_{\text{train}}^{\text{target}}$ represents the target future values for prediction.
Once trained, $M_{\text{base}}$ encapsulates a generalized representation of the time series dynamics but lacks specialization toward specific dominant patterns identified during validation.

\subsubsection{Transition from General to Specialized Weights}

The transition from a general DNN to pattern-specialized DNNs involves the fine-tuning of the base DNN $M_{\text{base}}$ on pattern-specific data. Each dominant pattern is represented by a cluster $C_k$, derived from the validation set $\mathcal{T}_{\text{val}}$. For each cluster $C_k$, we initialize a specialized DNN $M_k$ by copying the weights from the base DNN:
\begin{equation}
\theta_k^{(0)} \leftarrow \theta_{\text{base}}.
\end{equation}
This weight transfer ensures that each $M_k$ starts from a well-optimized initialization, benefiting from the knowledge encoded by $M_{\text{base}}$.
Subsequently, each $M_k$ undergoes fine-tuning on its respective cluster data $C_k$ to learn the characteristics of the corresponding pattern:
\begin{equation}
\theta_k = \arg\min_{\theta} \mathcal{L}_{\text{forecast}} \left( M_{\theta}(C_k), C_k^{\text{target}} \right).
\end{equation}
Fine-tuning on $C_k$ enables the DNN $M_k$ to adapt its parameters specifically to the regime represented by the cluster. The weights $\theta_k$ gradually shift from the global parameters $\theta_{\text{base}}$ to pattern-specialized parameters, capturing the nuances of localized temporal dependencies.

This two-stage training strategy leverages both the generalization capacity of the base DNN $M_{\text{base}}$ and the specialization capability of the pattern-specific DNNs $M_k$. Specifically, $M_{\text{base}}$ is trained on the entire historical dataset $\mathcal{T}_{\text{train}}$, enabling it to capture broad temporal features and general patterns inherent in the time series. Building on this foundation, each specialized DNN $M_k$ is further fine-tuned on a specific cluster of data $C_k$, allowing it to adapt to the unique temporal dynamics and regime-specific characteristics represented by that dominant pattern. This hierarchical approach ensures a balance between generalization and specialization, leading to models that are both robust and highly tailored to distinct patterns within the data. The transition of weights can be interpreted as:
\begin{equation}
\theta_k = \theta_{\text{base}} + \Delta_k,
\end{equation}
where $\Delta_k$ represents the adaptation from general dynamics to pattern-specific characteristics induced by $C_k$.

% \subsubsection{Benefits of Weight Transfer and Specialization}

% The proposed weight transfer and specialization strategy offers several benefits:
% \begin{itemize}
%     \item \textbf{Efficient convergence}: Since $\theta_{\text{base}}$ provides a well-initialized starting point, fewer updates are required to specialize $M_k$, reducing the risk of overfitting compared to training from scratch.
%     \item \textbf{Knowledge sharing}: The shared initialization preserves common temporal dependencies learned by $M_{\text{base}}$, while $\Delta_k$ focuses on divergences unique to $C_k$.
%     \item \textbf{Improved performance}: This adaptive strategy allows each $M_k$ to outperform $M_{\text{base}}$ on input sequences matching its cluster pattern.
% \end{itemize}

\subsection{Integration into Inference and Drift Adaptation}

During inference, the nearest dominant pattern to the most recently acquired pattern is identified via similarity to cluster centroids $\mu_k$, as described in Section~\ref{sec:cl}.  Given the current input window $S_t$ of size $p$, we compute its similarity to each cluster centroid $\mu_k$:
\begin{equation}
    k^* = \arg\min_k d(S_t, \mu_k)
\end{equation}
Once the most similar cluster is determined, its corresponding specialized DNN $M_{k^*}$ is deployed for forecasting.

In the presence of concept drift, new dominant patterns that are not represented by the existing clusters may emerge. To detect this, we monitor the evolution of the distance between the current pattern \(S_{t}\) and the closest cluster center: $d_{t} = \min_{k} d(S_{t}, \mu_k)$.
We record a reference distance \( d_{\text{ref}} \) at the start of the inference phase and continuously monitor deviations \(\delta_{t+h} = |d_{t+h}- d_{\text{ref}}|\). If the minimum distance diverges over time, i.e., meaning that the old pre-computed dominant patterns can not inform about the most recent patterns.
A drift is assumed to have occurred at $t+h$ if the true mean of $\delta$ significantly diverges from $0$. To determine this, we use the well-known Hoeffding-Bound~\cite{Hoeffding/63a}, which states that after $\omega$ independent observations of a real-value random variable with range $R$, its true mean has not diverged if the sample mean is contained within $\pm \xi$. Here, $\xi$ is defined as:
$\sqrt{\frac{R^2\ln(1/\gamma)}{2\omega}}$ 
with a probability of $1-\gamma$, a user-defined hyperparameter.
If a significant drift is detected, indicated by a divergence between the new input patterns and existing cluster centroids (e.g., exceeding a threshold $\delta$), we trigger the re-clustering process on a sliding window of recent data $\mathcal{T}_{\text{recent}}$.
%If a significant drift is detected, indicated by a divergence between the new input patterns and existing cluster centroids, we trigger the re-clustering process on a sliding window of recent data $\mathcal{T}_{\text{recent}}$.
New clusters $\mathcal{C}_{\text{new}} = \{ C'_k \}_{k=1}^{K'}$ and their centroids $\{ \mu'_k \}_{k=1}^{K'}$ are computed. To update the existing pool of clusters $\mathcal{C} = \{ C_k \}_{k=1}^{K}$, we evaluate the similarity between each new centroid $\mu'_j$ and the existing centroids $\mu_k$.
If a new cluster $C'_j$ represents a pattern that is sufficiently distinct from all existing clusters, it is incorporated as a new dominant pattern and a new specialized DNN $M'_k$ is fine-tuned on $C'_k$ and added to the pool of models. Formally, this is determined by:
%Formally, the update rule for cluster fusion is defined as:
\[
\text{If} \quad \min_{k} d(\mu'_j, \mu_k) > \tau, \quad \text{then add} \quad C'_j \quad \text{to} \quad \mathcal{C}
\]
where $d(\cdot,\cdot)$ denotes the distance metric (e.g., Euclidean distance) between the centroids, and $\tau$ is a predefined threshold controlling the similarity criterion for merging or adding clusters.
\section{Experiments}

\label{sec:experiments}

This work addresses the following key research questions to evaluate the effectiveness, adaptability, and efficiency of our proposed pattern-based fine-tuning framework for deep neural networks (DNNs) in time series forecasting:
\textbf{RQ1} How dominant patterns in time series data are identified and leveraged for model specialization?; \textbf{RQ2} Does pattern-based fine-tuning of DNNs yield improved forecasting accuracy compared to traditional training (Base)?; \textbf{RQ3} How does the integration of concept drift detection and online adaptation impact forecasting performance?;
\textbf{RQ4} What is the trade-off between forecasting accuracy and computational efficiency in the online drift-aware fine-tuning versus periodic blind adaptation?;
\textbf{RQ5} Can pattern-based fine-tuning compensate for suboptimal DNN architectures and improve performance under constrained model configurations?

\subsection{Experimental Set-up}

A total of $113$ real-world time series were utilized, originating from diverse application scenarios, including weather data, sensor readings, and financial forecasting. The datasets description is included in the supplementary material. The code and datasets are publicly available\footnote{\scriptsize\url{https://www.dropbox.com/scl/fo/2nipxobeqaeqx68mpvgj2/AIGsCNCuvFADR4JOHfvqwUk?rlkey=hj5a8sd6k6orwji6i4ptvgvov&st=p193wdwm&dl=0}}.
The list of all the parameters used in the Methodology is summarized in Table 
\ref{tab:pr}. 
\begin{table}[ht]
\centering
\caption{Summary of Parameters, Descriptions, and Values}
\resizebox{0.6\textwidth}{!}{\begin{tabular}{|l|l|l|}
\hline
\textbf{Parameter} & \textbf{Description} & \textbf{Value} \\
\hline
|$\mathcal{T}{\text{train}}$|, |\( \mathcal{T}{\text{val}} \)|,& Proportion of train, validation & 40\%, 40\%, 20\% \\
|\( \mathcal{T}{\text{test}} \)| &and test sets, respectively. &\\
$p$ & Length of the input subsequences used  & 10 \\
&for clustering and DNN training&\\
$K$ & Number of clusters & Auto. X-means \cite{pelleg2000x} \\
$n_{\text{min}}$&Minimum number of subsequences per cluster&10\% of |\( \mathcal{T}{\text{val}} \)| \\
$\gamma$ & The Hoeffding-Bound parameter & 0.05 \\
$\tau$ & Threshold for cluster fusion & 20\% of the Inter-Cluster Distance \\
\hline
\end{tabular}}
\label{tab:pr}
\end{table}
The methods used in the experiments were evaluated using the root mean squared error (RMSE). 
\paragraph{\textbf{Forecasting DNNs Set-up}}
To thoroughly assess the predictive capabilities across different DNN architectures, we considered a broad pool of DNNs that have been either traditionally employed for forecasting or adapted from other domains to meet forecasting challenges. The First family consists of classical deep learning models: Multi-Layer Perceptron (\textbf{MLP}), Long Short-Term Memory network (\textbf{LSTM)}, Convolutional Neural Networks (\textbf{CNN}), and a hybrid model combining CNN with an LSTM layer \textbf{(CNN-LSTM)}. 
The second group is composed of advanced deep learning architectures that have been specifically engineered to tackle the challenges of time series forecasting and are included in the GluonTS Python library \cite{alexandrov2020gluonts}: Temporal Regularized Matrix Factorization (\textbf{TRMF}), Long- and Short-term Time-series Network (\textbf{LSTNet}), Deep Global Local Forecaster (\textbf{DeepGlo}), \textbf{DeepState}, Deep Auto-Regressive (\textbf{DeepAR}), \textbf{DeepFactor}, \textbf{MQ-CNN}, and Temporal Fusion Transformer (\textbf{TFT}).

\paragraph{\textbf{Variants of Weights Learning and Fine-Tuning}}
For each DNN, we compute different weight learning or tuning versions:
\textbf{Base}: involves the traditional training where the weights are learned on the whole training time series data and kept static during inference;
The following fine-tuning versions are three variations of our pattern-based fine-tuning approach:
\begin{itemize}
    \item  \textbf{Offline-Tune}: After learning base weights, we apply our methodology for pattern-based fine-tuning. However, no adaptation using concept drift detection is included.
    \item  \textbf{Online-Tune}: Our full fine-tuning version using pattern-based specialization and concept drift adaptation;
\item  \textbf{Periodic-Tune}: Our fine-tuning version uses pattern-based specialization and blind adaptation in a periodic manner with each 10\% of upcoming time series observations of the new data in the test set.
\end{itemize}

\section{Results and Discussion}
\begin{figure}
    \centering
    \includegraphics[trim={0 0 0cm 0cm},clip,width=0.8\textwidth]{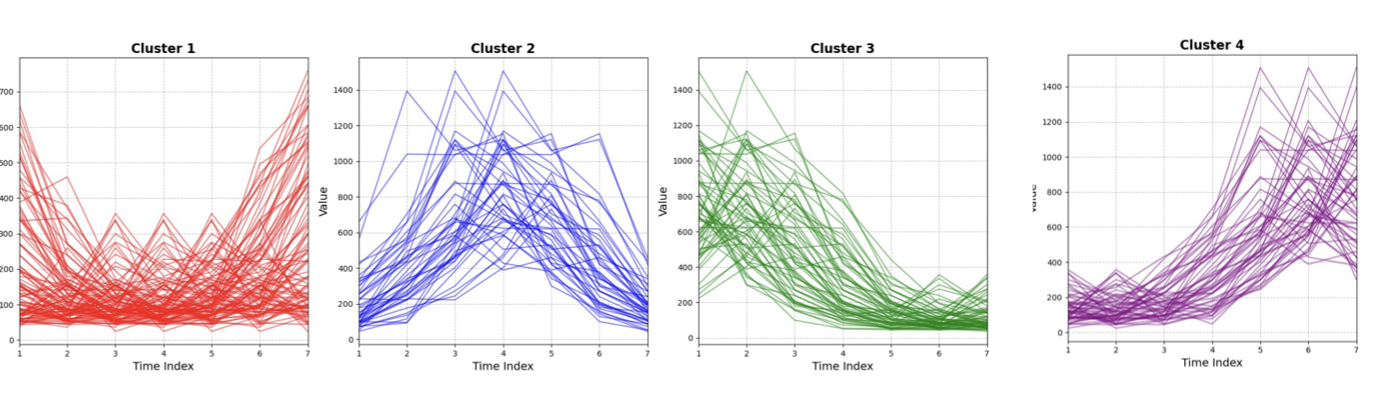}
    \caption{Clusters Visualization of the Tourism Time Series Data Set.}
    \label{fig:tr-cl}
\end{figure}

\begin{figure}
    \centering
    \includegraphics[trim={0 0 0cm 0cm},clip,width=0.9\textwidth]{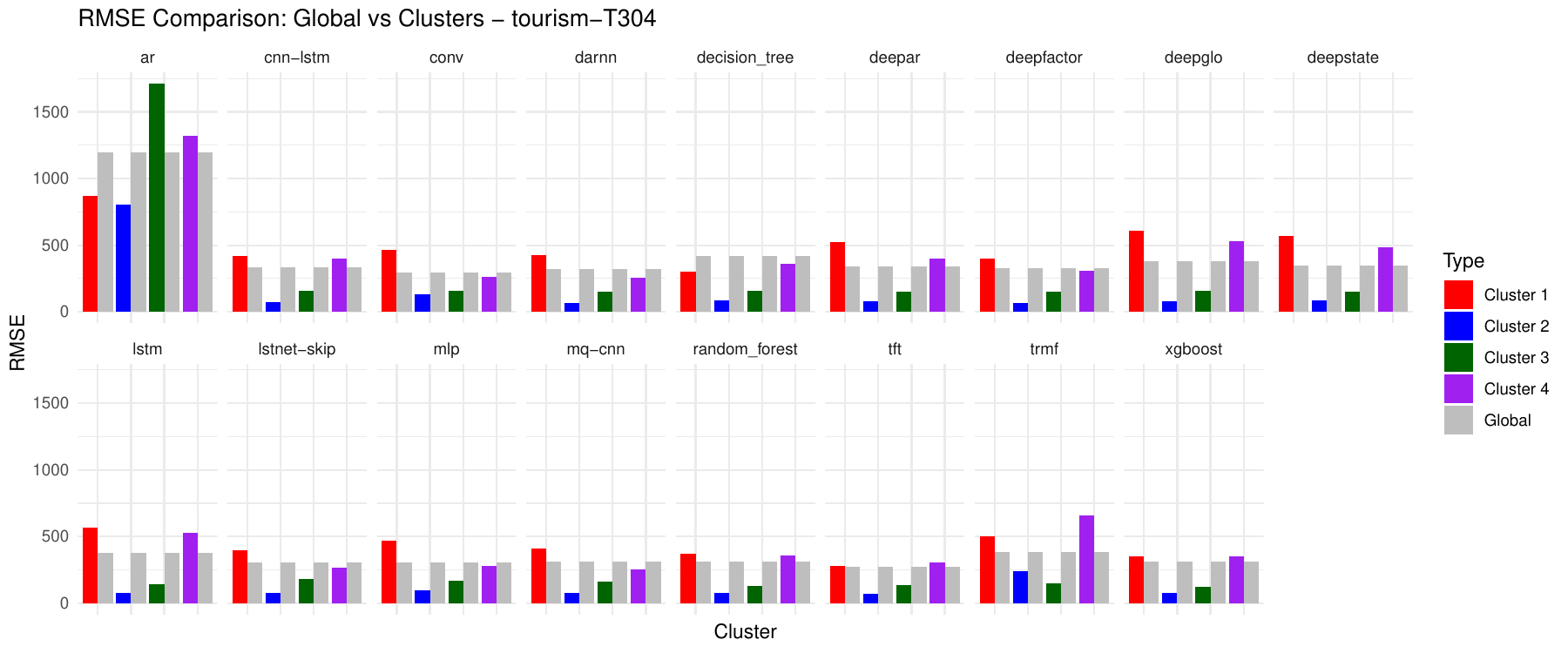}
    \caption{Global vs. Local RMSE Comparison on the Tourism Time Series Data.}
    \label{fig:rmse-cmp-t}
\end{figure}
\begin{figure}
    \centering
    \includegraphics[width=0.6\textwidth]{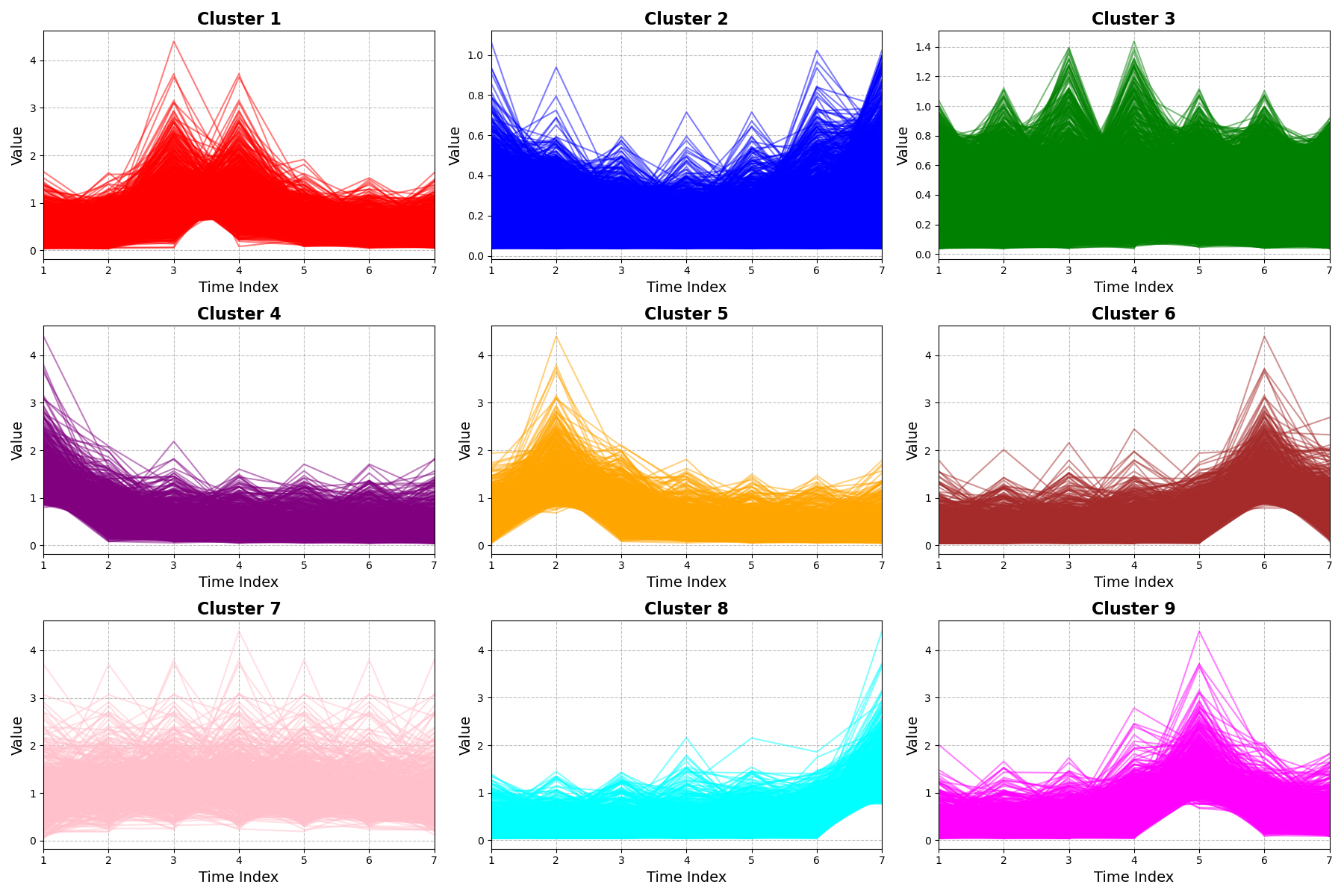}
    \caption{Clustered Subsequences Visualization of a time series in the London Smart Meters dataset}
    \label{fig:cl-london}
\end{figure}

Figures \ref{fig:cl-london} and \ref{fig:tr-cl} illustrate the results of clustering time series subsequences from the London Smart Meters and the Tourism datasets. Figure \ref{fig:rmse-cmp-t}  directly compares RMSE values per model across global and local forecasts (Clusters) on the Tourism data.
Table~\ref{tab:dnn_performance_comparison} presents a comprehensive comparison of the performance of different deep neural network (DNN) architectures under four distinct training and adaptation strategies: \textbf{Base}, \textbf{Offline-Tune}, \textbf{Online-Tune}, and \textbf{Periodic-Tune}. Each row corresponds to a specific DNN model (e.g., CNN, DeepGlo, CNN-LSTM, etc.) and reports its performance across several key metrics.
%including normalized Root Mean Square Error (nRMSE), standard deviation of nRMSE, average ranking across all datasets, and global ranking statistics.
The Avg. nRMSE/ Std. nRMSE columns represent the average and the standard deviation of the normalized root mean square error across all time series datasets. The Ranking DNN Arch. column represents the relative rank using the nRMSE of each architecture when averaged over all datasets for each family of DNN separately. %Avg. Global Ranking shows the model’s average global rank across all experiments. Std. Global Ranking shows the standard deviation in the models' global ranking, indicating how consistent the models' ranks are.

\begin{table*}[ht]
\centering
\caption{Performance comparison across different DNN architectures and tuning strategies.}
\label{tab:dnn_performance_comparison}
\resizebox{\textwidth}{!}{%
\begin{tabular}{|l|l|c|c|c|c|c|}
\hline
\textbf{Model} & \textbf{Strategy} & \textbf{Avg. nRMSE} & \textbf{Std. nRMSE} & \textbf{Ranking DNN Arch.} & \textbf{Avg. Global Ranking} & \textbf{Std. Global Ranking} \\
\hline
CNN & Base     & 0.166 & 0.17 & 3.70 & 29.47 & 12.61 \\
    & Offline  & 0.135 & 0.13 & 2.27 & 28.08 & 9.71 \\
    & \textbf{Online}  & \textbf{0.120} & \textbf{0.12} & 2.04 & 26.50 & 8.94 \\
    & \textbf{Periodic} & \textbf{0.113} & \textbf{0.10} & 1.39 & 20.80 & 8.33 \\
\hline
DeepGlo & Base     & 0.158 & 0.14 & 3.12 & 37.07 & 7.35 \\
        & Offline  & 0.144 & 0.13 & 2.49 & 34.22 & 8.65 \\
        & \textbf{Online}  & \textbf{0.131} & \textbf{0.12} & 2.16 & 33.22 & 9.61 \\
        & \textbf{Periodic} & \textbf{0.128} & \textbf{0.11} & 1.64 & 30.28 & 11.02 \\
\hline
CNN-LSTM & Base     & 0.158 & 0.15 & 3.70 & 31.73 & 9.44 \\
         & Offline  & 0.138 & 0.13 & 2.27 & 21.81 & 9.21 \\
         & \textbf{Online}  & \textbf{0.125} & \textbf{0.12} & 1.96 & 21.47 & 8.15 \\
         & \textbf{Periodic} & \textbf{0.118} & \textbf{0.11} & 1.46 & 20.64 & 9.08 \\
\hline
DeepFactor & Base     & 0.113 & 0.09 & 3.70 & 27.71 & 11.30 \\
           & Offline  & 0.095 & 0.08 & 2.32 & 15.20 & 10.62 \\
           & \textbf{Online}  & \textbf{0.091} & \textbf{0.08} & 2.05 & 16.23 & 10.79 \\
           & \textbf{Periodic} & \textbf{0.086} & \textbf{0.08} & 1.33 & 13.69 & 9.12 \\
\hline
DeepState & Base     & 0.136 & 0.11 & 3.69 & 31.73 & 9.18 \\
          & Offline  & 0.118 & 0.10 & 2.32 & 22.77 & 10.12 \\
          & \textbf{Online}  & \textbf{0.114} & \textbf{0.10} & 2.04 & 21.81 & 9.91 \\
          & \textbf{Periodic} & \textbf{0.104} & \textbf{0.09} & 1.35 & 15.86 & 8.48 \\
\hline
MQ-CNN & Base     & 0.168 & 0.17 & 3.63 & 29.79 & 13.41 \\
       & Offline  & 0.147 & 0.15 & 2.29 & 25.64 & 12.70 \\
       & \textbf{Online}  & \textbf{0.133} & \textbf{0.13} & 2.03 & 19.21 & 11.92 \\
       & \textbf{Periodic} & \textbf{0.124} & \textbf{0.12} & 1.45 & 19.93 & 9.84 \\
\hline
TFT & Base     & 0.115 & 0.10 & 3.55 & 26.84 & 13.72 \\
    & Offline  & 0.088 & 0.07 & 2.28 & 10.88 & 14.17 \\
    & \textbf{Online}  & \textbf{0.083} & \textbf{0.06} & 2.00 & 9.87 & 14.11 \\
    & \textbf{Periodic} & \textbf{0.079} & \textbf{0.06} & 1.57 & 9.12 & 14.03 \\
\hline
LSTM & Base     & 0.167 & 0.16 & 3.70 & 34.37 & 10.09 \\
     & Offline  & 0.153 & 0.14 & 2.26 & 26.33 & 11.62 \\
     & \textbf{Online}  & \textbf{0.141} & \textbf{0.14} & 2.00 & 25.42 & 11.00 \\
     & \textbf{Periodic} & \textbf{0.133} & \textbf{0.12} & 1.44 & 23.89 & 10.82 \\
\hline
MLP & Base     & 0.152 & 0.15 & 3.78 & 27.73 & 11.49 \\
    & Offline  & 0.127 & 0.12 & 2.17 & 16.26 & 8.23 \\
    & \textbf{Online}  & \textbf{0.112} & \textbf{0.11} & 1.98 & 15.31 & 8.04 \\
    & \textbf{Periodic} & \textbf{0.106} & \textbf{0.09} & 1.47 & 9.88 & 7.34 \\
\hline
LSTnet-Skip & Base     & 0.174 & 0.18 & 3.54 & 31.27 & 12.06 \\
            & Offline  & 0.123 & 0.12 & 2.32 & 19.20 & 10.91 \\
            & \textbf{Online}  & \textbf{0.105} & \textbf{0.10} & 2.05 & 12.65 & 10.24 \\
            & \textbf{Periodic} & \textbf{0.104} & \textbf{0.09} & 1.49 & 11.59 & 9.60 \\
\hline
DeepAR & Base     & 0.158 & 0.14 & 3.73 & 35.04 & 7.18 \\
       & Offline  & 0.142 & 0.13 & 2.32 & 27.63 & 7.96 \\
       & \textbf{Online}  & \textbf{0.129} & \textbf{0.12} & 2.05 & 26.67 & 7.80 \\
       & \textbf{Periodic} & \textbf{0.123} & \textbf{0.11} & 1.30 & 20.17 & 9.11 \\
\hline
\end{tabular}%
}
\end{table*}

Table~\ref{tab:runtime_comparison} compares the computational efficiency of the \textbf{Online-Tune} and \textbf{Periodic-Tune} strategies.
\begin{table}[ht]
\centering
\caption{Runtime Comparison Between Online-Tune and Periodic-Tune Strategies (in seconds). The table reports both the average runtime and its standard deviation across all datasets.}
\label{tab:runtime_comparison}
\resizebox{0.5\textwidth}{!}{\begin{tabular}{|l|c|c|}
\hline
\textbf{Model} & \textbf{Average Runtime (s)} & \textbf{Std Runtime (s)} \\
\hline
Online-Tune   & 15.62  & 11.15 \\
Periodic-Tune & 78.62 & 91.55 \\
\hline
\end{tabular}}
\end{table}
Table~\ref{tab:optimized_vs_underoptimized} presents a comparative analysis of the forecasting performance between optimized and under-optimized versions of various deep neural network (DNN) architectures. 
%The evaluation is based on the average normalized root mean square error (nRMSE), where lower values indicate superior forecasting accuracy.
The \textbf{Base (Opt)} and \textbf{Online (Opt)} columns correspond to the fully optimized models, trained and fine-tuned according to best practices. In contrast, the \textbf{Base (Under-Opt)} and \textbf{Online (Under-Opt)} columns reflect simplified architectures with reduced complexity, designed to assess the impact of our online fine-tuning methodology under suboptimal conditions.
\begin{table}[ht]
\centering
\caption{Comparison of Base and Online Fine-Tuning Performance Between Optimized and Under-Optimized Models. The values represent the average normalized RMSE (nRMSE) across datasets for each model configuration. Lower values indicate better forecasting performance.}
\label{tab:optimized_vs_underoptimized}
\resizebox{0.7\textwidth}{!}{\begin{tabular}{|l|c|c|c|c|}
\hline
\textbf{Model} & \textbf{Base (Opt)} & \textbf{Online (Opt)} & \textbf{Base (Under-Opt)} & \textbf{Online (Under-Opt)} \\
\hline
CNN          & 0.166 & 0.120 & 0.170 & 0.131 \\
CNN-LSTM     & 0.158 & 0.125 & 0.174 & 0.135 \\
DeepFactor    & 0.113 & 0.091 & 0.131 & 0.110 \\
DeepGlo       & 0.158 & 0.131& 0.166 & 0.150 \\
DeepState     & 0.136 & 0.114 & 0.166 & 0.146\\
LSTM          & 0.167 & 0.141 & 0.177 & 0.150 \\
LSTNet-Skip   & 0.174 & 0.105 & 0.181 & 0.125 \\
MQ-CNN        & 0.168 & 0.133 & 0.170 & 0.139 \\
TFT           & 0.115 & 0.083 & 0.148 & 0.119 \\
\hline
\end{tabular}}
\end{table}
\subsubsection{RQ1: Dominant Patterns in Time Series Data}
The identification of dominant patterns in time series data is a cornerstone of our proposed fine-tuning framework. We show two representative examples of time series datasets, namely London energy consumption and Tourism demand forecasting.

For the Tourism time series data in Figure \ref{fig:tr-cl}, the clustering process segments the data into regimes with varying complexity. Cluster 1 captures noisy patterns with frequent local fluctuations, posing significant challenges for accurate forecasting. In contrast, Clusters 2, 3, and 4 demonstrate well-defined trend structures—Cluster 2 follows a rise-and-fall trajectory, and Cluster 3 shows a consistent downward trend. These clearly structured patterns are more conducive to learning and forecasting by DNN models. Cluster 4, characterized by an increasing trend with high variability, highlights the challenge of forecasting under volatile conditions. As illustrated in Figure \ref{fig:tr-cl}, the localized forecasting performance of models, when specialized for specific clusters, varies from global models trained on the entire time series data. This confirms that dominant pattern identification through clustering can effectively support localized forecasting strategies by decomposing complex time series into simpler, learnable regimes.

Similarly, for the London time series data, the clustering results, illustrated in Figure \ref{fig:cl-london}, reveal distinct and well-structured patterns. Each subplot corresponds to a cluster, where the grouped subsequences exhibit highly similar temporal behaviors. Specifically, Clusters 1 and 5 exhibit sharp peaks followed by gradual declines, which correspond to recurring peaks in energy consumption during certain periods (e.g., daily or seasonal usage spikes). Cluster 7 displays regular periodic fluctuations, indicative of cyclical energy demand, while Clusters 2 and 3 show patterns characterized by relatively low variability with localized peaks. This segmentation indicates that the clustering method effectively differentiates between various temporal regimes, enabling the specialization of models on distinct patterns.

\subsubsection{RQ2: Comparison of Pattern-based Fine-Tuning of DNNs to Traditional Training (Base)}
The results presented in Table \ref{tab:dnn_performance_comparison} provide strong empirical evidence that pattern-based fine-tuning significantly enhances forecasting accuracy across a wide range of DNN architectures. Specifically, our Offline-Tune strategy, which fine-tunes DNNs on dominant patterns without incorporating online adaptation, consistently and significantly outperforms the Base approach. %On average, the Offline-Tune strategy reduces the normalized RMSE (nRMSE) by approximately 15%-20% compared to the base models. 
The introduction of the adaptation to new emerging patterns, either blindly with the Periodic-Tune version or in an informed manner following concept-drift detection with Online-Tune, further improved the forecasting accuracy compared to the traditional base approach.
For example, the CNN-LSTM architecture exhibits a reduction in nRMSE from 0.158 (Base) to 0.138 (Offline-Tune), while the LSTnet-Skip model improves from 0.174 to 0.123. Further improvement is achieved with the Online and Periodic-Tune to reach 0.105 and 0.104, respectively. These improvements are attributed to the enhanced ability of the fine-tuned models to specialize in the localized temporal dynamics of distinct clusters, as opposed to relying on a generalized model trained on heterogeneous data.

Furthermore, the consistent performance gains observed across different architectures—including both RNN-based models (e.g., LSTM, DeepAR) and attention-based models (e.g., TFT)—highlight the general applicability and robustness of our pattern-based fine-tuning strategy. This is confirmed by the local average ranking per architecture that shows consistently lower ranks (better performance, rank 1 means the method is the best on all the data sets) for the Periodic and the Online versions. 
The global ranking across all the models and the datasets (the last two columns in Table \ref{tab:dnn_performance_comparison}) confirms these findings. It is clear that the  Offline, Periodic, and Online have significantly lower ranks compared to the Base version. For example, for MLP, while the base traditional training scores an average rank of 27.73, the rank drops significantly to lower than 17 with the pattern-based fine-tuning versions. Similar to DeepFactor, we notice a decrease in the global average rank from 27.71 to 16.23 and 13.69 with the Periodic and the Online-Tune, respectively.

The results confirm that by leveraging dominant patterns identified during validation, our method facilitates better model adaptation to specific temporal structures, improving overall forecasting accuracy.

\subsubsection{RQ3/RQ4:Trade-off Forecasting Accuracy and Computational Efficiency in Online Adapatation} 
The Online-Tune strategy extends the Offline-Tune approach by integrating concept drift detection to adapt to emerging patterns during inference. As shown in Table \ref{tab:dnn_performance_comparison}, Online-Tune yields further performance improvements over Offline-Tune across all DNN architectures.
For instance, LSTnet-skip improves its nRMSE from 0.123 (Offline) to 0.105 (Online), and TFT achieves an improvement from 0.088 to 0.083. The average global rankings also improve under the Online-Tune strategy, indicating its ability to maintain superior performance over time, even as data distributions shift.
The dynamic adaptation mechanism, driven by concept drift detection, enables the recalibration of model weights and the introduction of new specialized models for newly detected patterns. This prevents performance degradation due to non-stationarity in the data and ensures that the models remain aligned with the evolving temporal dynamics. 

However, when compared to the Periodic-Tune strategy, Online-Tune generally yields lower accuracy. Specifically, the Periodic-Tune approach achieves the best overall Avg. nRMSE across the majority of models (e.g., CNN: 0.113, DeepFactor: 0.086, and TFT: 0.079), outperforming Online-Tune.
This performance gap can be attributed to several potential factors. First, in the Online-Tune strategy, the reliance on concept drift detection mechanisms may lead to missed drift events, particularly when the detection sensitivity is not optimally configured. As a result, the system may fail to recognize subtle or gradual changes in the data distribution, leading to delayed or insufficient model updates. Second, the drift detection mechanism may introduce latency, where adaptation occurs after a drift has already affected forecasting accuracy. 
%Finally, false negatives in drift detection can prevent necessary retraining steps, while false positives can result in unnecessary adaptation, introducing instability.
In contrast, the Periodic-Tune strategy blindly triggers adaptation at regular intervals (every 10\% of new observations in our setting), ensuring continuous updates irrespective of drift signals. This proactive behavior, though computationally more expensive, guarantees that the model periodically realigns with the latest data distribution, leading to superior accuracy at the cost of a significantly higher runtime. The average runtime of Online-Tune is 15.62 seconds, compared to 78.62 seconds for Periodic-Tune, representing an 80\% reduction in computational overhead.
%This efficiency is attributed to the targeted nature of the online adaptation process, which selectively updates models in response to detected concept drifts, as opposed to Periodic-Tune, which blindly retrains models at fixed intervals, regardless of whether the data distribution has changed.

%(Table~\ref{tab:runtime_comparison} shows that Periodic-Tune runtime is significantly higher than Online-Tune).

To close the performance gap between Online-Tune and Periodic-Tune, further tuning of the concept drift detection parameters is warranted. Additionally, combining drift detection with auxiliary performance monitors (e.g., tracking local RMSE or coverage metrics) could improve the robustness of the Online-Tune strategy. 
These enhancements will allow for more precise, timely adaptations while preserving the computational efficiency demonstrated by the current implementation.

% \subsubsection{RQ4:Trade-off Forecasting Accuracy and Computational Efficiency }
% While Online-Tune offers substantial improvements in forecasting performance, it also aims to balance computational efficiency. Table \ref{tab:runtime_comparison} compares the runtime of Online-Tune and Periodic-Tune, revealing that Online-Tune is significantly more efficient.
% The average runtime of Online-Tune is 4.62 seconds, compared to 78.62 seconds for Periodic-Tune, representing a 94\% reduction in computational overhead. This efficiency is attributed to the targeted nature of the online adaptation process, which selectively updates models in response to detected concept drifts, as opposed to Periodic-Tune, which blindly retrains models at fixed intervals, regardless of whether the data distribution has changed.

% Despite its lower computational cost, Online-Tune improves the forecasting accuracy over the Offline-Tune in all cases. For example, while Periodic-Tune marginally outperforms Online-Tune in certain cases (e.g., DeepAR nRMSE of 0.12 vs. 0.14), the overall trade-off favors Online-Tune, particularly when resource constraints or real-time processing requirements are considerations. These findings demonstrate that Online-Tune offers an effective balance between accuracy and efficiency, making it well-suited for deployment in operational forecasting systems.

\subsubsection{RQ5: Compensation for Suboptimal DNN Architectures using Pattern-based Fine-tuning}
The results presented in Table~\ref{tab:optimized_vs_underoptimized} offer compelling evidence regarding the capacity of our pattern-based fine-tuning framework to mitigate the limitations of under-optimized deep neural network (DNN) architectures. Specifically, we compare forecasting performance across two axes: optimized versus under-optimized models and standard Base training versus our Online fine-tuning approach.

For all tested DNN models, under-optimized architectures—characterized by simplified structures and reduced parameter complexity—predictably yield inferior forecasting performance when trained with conventional methods (Base Under-Opt). For example, the CNN-LSTM model's base performance degrades from an average normalized RMSE (nRMSE) of 0.158 (Base Opt) to 0.174 (Base Under-Opt). Similar degradations are observed for DeepState (0.136 to 0.166) and TFT (0.115 to 0.148).
However, once our pattern-based Online Fine-Tuning is applied, the under-optimized models show substantial improvements. In several cases, the performance of these simpler models approaches or even exceeds that of their fully optimized counterparts under standard training. For instance, the CNN model improves from 0.170 (Base Under-Opt) to 0.131 (Online Under-Opt), exceeding its optimized Base version (0.166) and closely matching its optimized Online performance (0.120). Despite being under-optimized, the DeepFactor model achieves an nRMSE of 0.110 with online fine-tuning, narrowing the gap with its optimized version (0.091). MQ-CNN shows a marked improvement from 0.170 (Base Under-Opt) to 0.139 (Online Under-Opt), closing in on its optimized Online performance (0.133).
These findings indicate that our approach is effective in compensating for reduced architectural capacity by leveraging specialized adaptation to dominant temporal patterns. The fine-tuning process enables even simplified networks to focus on pattern-specific dynamics, extracting relevant information with fewer parameters and improving generalization within localized regimes.
The practical implications are significant in resource-constrained environments—such as edge devices or real-time applications—and our method allows the deployment of lighter DNN architectures without sacrificing substantial accuracy. Furthermore, this compensatory effect reduces the dependency on complex and computationally expensive hyperparameter optimization and architectural search procedures.

%In summary, the results demonstrate that pattern-based fine-tuning provides a robust mechanism for enhancing the forecasting capability of under-optimized DNNs. By tailoring models to specific temporal regimes and continuously adapting through online updates, our approach achieves a favorable balance between model complexity and predictive performance.

%revealing distinct and well-structured patterns. Each subplot represents a distinct cluster, where the grouped subsequences exhibit similar structural characteristics. The clustering process successfully captures variations in consumption patterns, with certain clusters displaying pronounced peaks at specific time indices, while others exhibit smoother trends or lower fluctuations. For example, Clusters 1 and 5 exhibit a sharp peak followed by a gradual decline, indicating a recurring pattern in energy consumption. Clusters 3 and 6 display periodic fluctuations, suggesting cyclical consumption behavior.Meanwhile, Clusters 2 and 8 capture patterns with lower overall variability but clear localized peaks. This clear segmentation suggests that the clustering methodology effectively distinguishes different temporal structures.

\section{Concluding Remarks and Future Work}
\label{sec:conclusion}
This paper introduced a novel framework for pattern-based fine-tuning of deep neural networks (DNNs) in time series forecasting. By identifying dominant temporal patterns through clustering and fine-tuning specialized DNNs on these distinct regimes, the proposed approach enables models to adapt effectively to the dynamic and non-stationary characteristics of real-world time series. Extensive experiments demonstrated consistent gains in forecasting accuracy across both optimized and under-optimized DNN architectures. Additionally, we highlighted the trade-off between online and periodic fine-tuning in terms of accuracy and computational efficiency.
Future work will focus on enhancing the concept drift detection mechanism by adaptively tuning its sensitivity parameters to reduce missed drifts. We also aim to explore advanced dynamic clustering methods for more robust pattern identification. Integrating automated architecture search and hyperparameter optimization will be considered to further improve the adaptability and efficiency of the proposed framework.

\bibliographystyle{splncs04}
\bibliography{mybib}

\end{document}